\pdfoutput=1

\documentclass[runningheads]{llncs}
\usepackage{graphicx}
\usepackage{comment}
\usepackage{amsmath,amssymb} 
\usepackage{color}

\usepackage{multirow}
\usepackage{booktabs}
\usepackage{makecell}
\usepackage{subfigure}

\usepackage{hyperref}
\usepackage[perpage]{footmisc}

\begin{document}
	\pagestyle{headings}
	\mainmatter
	\def\ECCVSubNumber{520}  
	
	\title{Social Adaptive Module for Weakly-supervised Group Activity Recognition} 

	\titlerunning{Social Adaptive Module}
	%
	\author{Rui Yan\inst{1} \and
		Lingxi Xie\inst{2} \and
		Jinhui Tang\inst{1}\thanks{Corresponding author} \and
		Xiangbo Shu\inst{1} \and
		Qi Tian\inst{2}}
	\authorrunning{R. Yan {\em et al.}}
	%
	\institute{School of Computer Science and Engineering, Nanjing University of Science and Technology, Nanjing, China \and
		Huawei Inc., China\\
		\email{\{ruiyan, jinhuitang, shuxb\}@njust.edu.cn, 198808xc@gmail.com, tian.qi1@huawei.com}}
	\maketitle
	
	\begin{abstract}
		{This paper presents a new task named weakly-supervised group activity recognition (GAR) which differs from conventional GAR tasks in that only video-level labels are available, yet the important persons within each frame are not provided even in the training data. This eases us to collect and annotate a large-scale NBA dataset and thus raise new challenges to GAR. To mine useful information from weak supervision, we present a key insight that key instances are likely to be related to each other, and thus design a social adaptive module (SAM) to reason about key persons and frames from noisy data. Experiments show significant improvement on the NBA dataset as well as the popular volleyball dataset. In particular, our model trained on video-level annotation achieves comparable accuracy to prior algorithms which required strong labels.}
		
	\end{abstract}

	\section{Introduction}
	Group activity recognition~(GAR) has a variety of applications in video understanding, such as sports analysis, video surveillance, and public security. Compared with traditional individual actions~\cite{long2018attention,soomro2012ucf101,Kuehne11,gan2015devnet,lin2019tsm}, group activities~({\em a.k.a}, collective activities)~\cite{choi2009they,ibrahim2016hierarchical,yan2018participation,wang2017recurrent} are performed by multiple persons cooperating with each other. Thus, the models for GAR require to understand not only the individual behaviors but also the relationship between each person.
	
	Previous fully-supervised methods which require person-level annotation~({\em i.e.} ground-truth bounding boxes and individual action label for each person, even interaction label for person-person pairs) have achieved promising performance on group activity recognition. Typically, these methods~\cite{ibrahim2016hierarchical,yan2018participation,tang2018mining,tang2019coherence,shu2017cern,wang2017recurrent,azar2019convolutional,bagautdinov2017social,wu2019learning} extract feature for each people according to the corresponding bounding boxes supervised by individual action label, and then fuse person-level feature into a single representation for each frame. {However, previous methods are sensitive to the varying number of people in each frame and require the explicit locations of them, which is limited in practical applications.}
	
\begin{figure}[!t]
		\centering
		\includegraphics[width=0.95\textwidth]{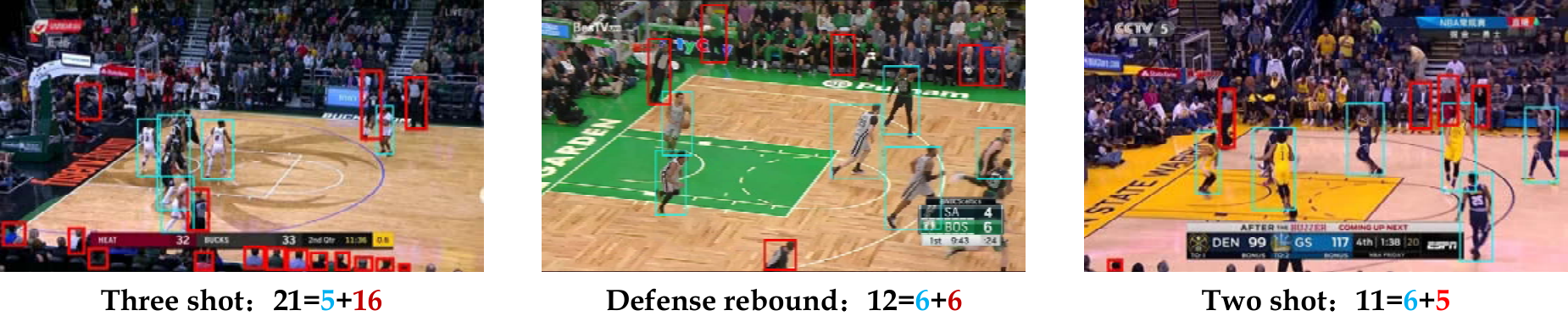}
		\caption{{\it{Best viewed in color}}. Illustration of the uncertain input issue under weakly-supervised setting. For different activities, the off-the-shelf detector will generate varying numbers of proposals, most of which~(in {\color{red}red} boxes) are useless for recognizing group activities. For instance, ``{\bf Three shot:} $\bf21$ = {\color{cyan} $\bf5$} + {\color{red}$\bf16$}" means that the detector generates a total of $\bf21$ proposals, but only {\color{cyan}$\bf5$} of them are players and other {\color{red}$\bf16$} proposals are outliers in an activity of three-shot}
		\label{fig:Motivation}
	\end{figure}

	To this end, we investigate GAR in a weakly-supervised setting which only provides video-level labels for each video clip. This setting not only is practical to real-world scenarios but also provides a simpler and lower-cost way for the annotation of new benchmarks. Benefiting from it, we collect a larger and more challenging benchmark, NBA, consisting of $181$ basketball games which involve more long-term temporal and fast-moving activities. Meanwhile, the weakly-supervised setting also brings {uncertain input} issue in each frame, as illustrated in Fig.~\ref{fig:Motivation}. Under this setting, lots of useless proposals will be fed into the approach. Besides, numerous irrelevant frames will also appear in the video clip, if the temporal structure of activities~({\em e.g.}, in NBA) is long.

	To tackle these issues, we further propose a simple yet effective module, namely Social Adaptive Module~(SAM), which can adaptively select discriminative proposals and frames from the video for weakly-supervised GAR. SAM aims at assisting the weakly-supervised training by leveraging a social assumption that {\bf key instances~(people/frames) are highly related to each other}. Specifically, we firstly construct a dense relation graph on all possible input feature to measure the relatedness between each other, then pick the top ones according to their relatedness. Based on the selected feature, a sparse relation graph is built to perform relational embedding for them. Benefiting from SAM, our approach trained without fully-supervision still obtains the comparable performance to previous methods on the popular volleyball dataset~\cite{ibrahim2016hierarchical}. 
	
	{Our contributions include: (a) The weakly-supervised setting that only provides video-level labels is introduced for GAR. (b) Thanks to this setting, a larger and more challenging benchmark, {NBA}, is collected from the web at a low cost. (c) To ease the weakly-supervised training, a SAM is proposed to adaptively find the effective person-level and frame-level representation based on the social assumption that key instances are usually closely related to each other.}

	\section{Related Work}
	{\bf Group Activity Recognition.}~Initial approaches~\cite{ibrahim2016hierarchical,yan2018participation,tang2018mining,wang2017recurrent} for recognizing group activities adopted the two-stage pipeline. They pre-extracted feature for each person from a set of patch images and then fuse them into a single vector for each frame by various methods~({\em e.g.}, pooling strategies~\cite{ibrahim2016hierarchical,shu2017cern}, attention mechanism~\cite{qi2018stagnet,tang2018mining,yan2018participation}, recurrent models~\cite{deng2016structure,tang2019coherence,wang2017recurrent,yan2018participation}, graphical models~\cite{amer2014HiRF,lan2012discriminative,lan2012social}, and AND-OR grammar models~\cite{amer2012cost,shu2015joint}). Nevertheless, these two-stage methods separate feature aggregation from representation learning, which is not conducive to a deep understanding of group activities. 
	To this end, Bagautdinov~\emph{et al.}~\cite{bagautdinov2017social} introduced an end-to-end framework to jointly detect multiple individuals, infer their individual actions, and estimate the group activity. Wu~\emph{et al.}~\cite{wu2019learning} extended~\cite{bagautdinov2017social} by stacking multiple graph convolutional layer to infer the latent relation between each person. Azar~\emph{et al.}~\cite{azar2019convolutional} constructed an activity map based on bounding boxes and explore the spatial relationships among people by iteratively refining the map. However, all of the above methods still require the action-level supervision~(action labels and bounding boxes for each person), which is time-consuming to tag. {Ramanathan~\emph{et al.}~\cite{ramanathan2016detecting} detected events and key actors in multi-person videos without individual action labels, but they still needed to annotate the bounding boxes of all the players in a subset of $9,000$ frames for training a detector.} This work introduces a more practical weakly-supervised setting that only provides video-level labels for group activity recognition.

	{\bf Existing Datasets Related to GAR.}~Limited by the time-consuming tagging, there are currently only four datasets for understanding group activities, as shown in Table~\ref{table:existing_datasets}. Choi~\emph{et al.}~\cite{choi2009they} proposed the first dataset, Collective Activity Dataset~(CAD), consisting of real-world pedestrian sequences. Then, Choi~\emph{et al.}~\cite{choi2011learning} extended CAD to CAED by adding two new actions~({\em i.e.}, ``Dancing" and ``Jogging") and removing the ill-defined action~({\em i.e.}, ``Walking"). There is no specific group activity defined in CAD and CAED, in which the scenarios are assigned group activities based on majority voting. Moreover, Choi and Savarese~\cite{choi2012unified} collected a Choi's New Dataset~(CND) composed of many artificial pedestrian sequences. Recently, Ibrahim~\emph{et al.}~\cite{ibrahim2016hierarchical} introduced a sports video dataset, Volleyball Dataset~(VD), which contains numerous volleyball games. However, as the largest and most popular dataset, VD contains quite a few wrong labels which directly affect the evaluation of proposed approaches. 
	{In addition, Ramanathan~\emph{et al.}~\cite{ramanathan2016detecting} released NCAA but few researchers have used it for GAR since only YouTube video links are provided and many of them are dead now. Some activities~({\em e.g.}, “steal”, “slam dunk *” and “free-throw *”) in NCAA can be recognized using one key frame, which actually evades from some key challenges of GAR.} Limited by the size and quality of the above datasets, the recent studies of group activity recognition have encountered the bottleneck. In this work, we collect a larger and more challenging dataset from the basketball games and do not provide any person-level information~({\em i.e.}, the bounding boxes and action labels for each person), thanks to the weakly-supervised setting. Moreover, compared with previous benchmarks, our NBA contains more activities that involve long-term temporal structure and are fast-moving.
	
	{\bf Relational Reasoning.}~Recently, relationships among entities~({\em i.e.}, pixels, objects or persons) have been widely leveraged in various computer vision tasks, such as Visual Question Answering~\cite{santoro2017simple,jang2017tgif,cadene2019murel}, Scene Graph Generation~\cite{johnson2015image,li2017scene,yang2018graph}, Object Detection~\cite{hu2018relation,chen2017spatial}, and Video Understanding~\cite{zhou2018temporal,wang2018non,liu2019learning}. Santoro~\emph{et al.}~\cite{santoro2017simple} presented a relational network module to infer the potential relationships among objects for improving the performance of visual question answering.  Hu~\emph{et al.}~\cite{hu2018relation} embedded a relation module into existing object detection systems for simultaneously detecting a set of objects and interactions between their appearance and geometry. Besides the spatial relationship among objects in the image, some recent works also explored the temporal relational structure of the video. Liu~\emph{et al.}~\cite{liu2019learning} proposed a novel neural network to learn video representations by capturing potential correspondences for each feature point. Moreover, some recent methods~\cite{deng2016structure,qi2018stagnet,wu2019learning} explored the spatial relationships between each people in group activities. In this work, we apply relational reasoning to choose the most relevant people from a number of proposals for weakly-supervised GAR.

	\section{Weakly-supervised Group Activity Recognition}\label{weakly-supervised_GAR}
	\subsection{Weakly-supervised Setting}
	{For a more practical group activity recognition, i) the number of people in the scene varies over different activities even time, and ii) the person-level annotations cannot be provided in real-world applications. Therefore, we introduce a weakly-supervised setting that {\em only video-level labels are available, yet the location and action label of each person are not provided}.
		
		In this work, the task of recognizing group activity under this setting is called weakly-supervised GAR which aims to directly recognize the activity performed by multiple collectively from the video with only a video-level label during training. Apparently, weakly-supervised GAR can be applied to more complex and real-world applications~({\em e.g.}, real-time sports analysis and video surveillance) which cannot provide fine-grained supervision. Besides, the weakly-supervised setting eases the annotation of benchmarks for the task. Without annotating the person-level supervision, we only require {$\frac{1}{2K+1}$} tagging labor$\footnote{The fully-supervised setting requires $K$ boxes, $K$ actions, and $1$ group activity, but the weakly-supervised setting only needs 1 group activity label. We roughly assumed the same labor for each annotation.}$ as before where $K$ is the number of people in the scene.
	}

	\subsection{The NBA Dataset for Weakly-supervised GAR}\label{NBA}
	{Under the weakly-supervised setting, we introduce a new video-based dataset, the NBA dataset. It describes the group activities that are common in basketball games. There is no annotation for each person and only a group activity label assigned to each clip. To the best of our knowledge, it is currently the largest and most challenging benchmark for group activity analysis, as shown in Table.~\ref{table:existing_datasets}. We will introduce the NBA dataset from the following aspects: the source of the video data, the effective annotation strategy, and the statistics of this dataset.}
	
	\setlength{\tabcolsep}{4pt}
	\begin{table}[t]
		\begin{center}
			\caption{Comparison of the existing datasets for group activity recognition}
			\label{table:existing_datasets}
			\begin{tabular}{lccccccc}
				\hline\noalign{\smallskip}
				Dataset &\# Videos &\# Clips & \makecell[c]{\# Individual \\ Actions}& \makecell[c]{\# Group \\ Activities} &\makecell[c]{Activity \\ Speed} &\makecell[c]{Camera \\ Moving} \\
				\noalign{\smallskip}
				\hline
				\noalign{\smallskip}
				CAD~\cite{choi2009they}      &44         & $\approx$ $2,500$             &5            & 5            &  slow            &N\\
				CAED~\cite{choi2011learning}     &30            & $\approx$ $3,300$            &6                & 6            & slow            &N\\
				CND~\cite{choi2012unified}     &32            &  $\approx$ $2,000$        &3                &6            &slow        &N\\
				VD~\cite{ibrahim2016hierarchical}  &55    &$4,830$    &9        & 8     &medium        &Y\\
				{\bf NBA}~(ours)                         &181    &$9,172$        &-        & 9     & fast            &Y\\
				\hline
			\end{tabular}
		\end{center}
	\end{table}
	\setlength{\tabcolsep}{1.4pt}
	
	{\bf Data Source.} It is a natural choice to collect videos of team sports for studying group activity recognition. In this work, we collect a subset of the 181 NBA games of 2019 periods from the web. Compared with the activities in volleyball games~\cite{ibrahim2016hierarchical}, the ones in basketball games have more long-term temporal structure and fast moving-speed, which brings up new challenges to group activity analysis. For one thing, the number of players may vary over different frames. On the other hand, the activity is so fast that the single-frame based person-level annotation is useless to track these players. Therefore, it is difficult to label all people in these videos which differs from volleyball games, thus we annotate this benchmark under the weakly-supervised setting. Due to the copyright restriction, this dataset is available upon request.
	
	{\bf Annotation.} Given a video, the goal of annotation is to assign the group activities to the corresponding segments. It is time-consuming to manually label such a huge dataset with conventional annotation tools. To improve the annotation efficiency, we take full advantage of the logs provided by the NBA's official website and design a simple and automatic pipeline to label our dataset. There are three steps: i) Filter out some unwanted records in the log file corresponding with a video. ii) Identify the timer in each frame by Tesseract-OCR~\cite{smith2007overview} and match it with the valid records generated from step i. iii) Save the segments with a fixed length according to the time points obtained from step ii.
	
	{\bf Statistics.} We collect a total of 181 videos with a high resolution of $1920 \times 1080$. Then we divide each video into 6-second clips by the above-mentioned annotation method and sub-sample them to 12fps.
	Besides, we remove some abnormal clips which contain close-up shots of players or instant replays. Ultimately, there are a total of $9,172$ video clips, each of which belongs to one of the 9 activities. Here, we drop some activities such as ``dunk" and ``turnover" due to the limited sample size, and do not use ``free-throw" that is easy to be distinguished. We randomly select $7,624$ clips for training and $1,548$ clips for testing. Table~\ref{table:Statistics} shows the sample distributions across different categories of group activities and the corresponding average number of people in the scene.
	
	\setlength{\tabcolsep}{4pt}
	\begin{table}[!t]
		\begin{center}
			\caption{Statistics of the group activity labels in NBA. ``2p", ``3p", ``succ", ``fail", ``def" and ``off" are abbreviations of ``two points", ``three points", ``success", ``failure", ``defensive rebound" and ``offensive rebound", respectively}
			\label{table:Statistics}
			\begin{tabular}{lcccccccccc}
				\hline\noalign{\smallskip}
				\multicolumn{2}{l}{Group Activity} &\makecell[c]{2p\\-succ.} &\makecell[c]{2p\\-fail.\\-off.} &\makecell[c]{2p\\-fail.\\-def.}  &\makecell[c]{2p\\-layup\\-succ.} &\makecell[c]{2p\\-layup\\-fail.\\-off.} &\makecell[c]{2p\\-layup\\-fail.\\-def.}  &\makecell[c]{3p\\-succ.} &\makecell[c]{3p\\-fail.\\-off.} &\makecell[c]{3p\\-fail.\\-def.}\\
				\noalign{\smallskip}
				\hline
				\noalign{\smallskip}
				\multirow{2}{*}{\# clips}&{\makecell[c]{Train}} &\makecell[c]{798} &\makecell[c]{434} &\makecell[c]{1316} &\makecell[c]{822} &\makecell[c]{455} &\makecell[c]{702} &\makecell[c]{728} & \makecell[c]{519} &\makecell[c]{1850}\\
				\cmidrule(lr){2-11}
				&Test &\makecell[c]{163} &\makecell[c]{107} &\makecell[c]{234} &\makecell[c]{172} &\makecell[c]{89} &\makecell[c]{157} &\makecell[c]{183} & \makecell[c]{83} &\makecell[c]{360}\\
				\hline
			\end{tabular}
		\end{center}
	\end{table}
	\setlength{\tabcolsep}{1.4pt}
	
	\section{Approach}\label{Approach}
	\subsection{Mining Key Instances via Social Relationship}\label{Overall_Formulation}
	{In general, the key and difficult point in obtaining category information from visual input is to construct and learn their intermediate representation. For the task of group activity recognition, such intermediate representation made up of individual feature and underlying relationships among them, refers to {\em social-representation} in this paper. The previous fully-supervised setting~\cite{ibrahim2016hierarchical,wang2017recurrent,yan2018participation} provides a variety of extra fine-grained supervision information~({\em e.g.}, ground-truth bounding box and action label for each person, and even the interaction label for each person-person pair) to ensure that social-representation can be constructed and learned stably during training. However, under the weakly-supervised setting which only provides video-level labels, it is difficult for models to define and learn discriminative social-representation stably.}
	
\begin{figure}[!t]
		\centering
		\includegraphics[width=0.9\textwidth]{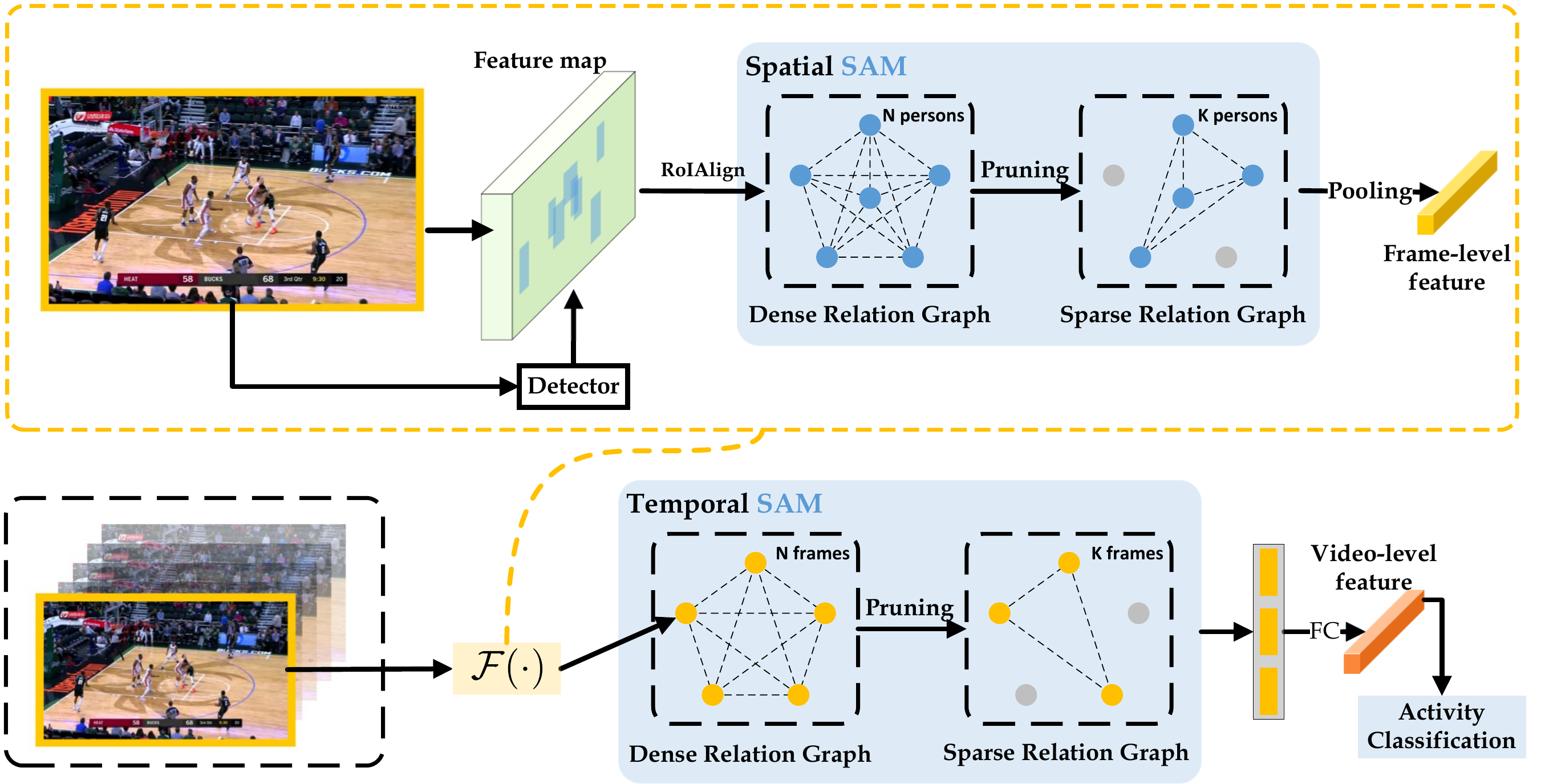}
		\caption{Overview of our approach for weakly-supervised GAR. The inputs are a set of frames and the associating pre-detected bounding boxes for people. We apply SAM to concurrently select discriminative person-level feature in spatial domain and effective frame-level representations in temporal domain~(Best viewed in color)}
		\label{fig:Overview}
	\end{figure}

	To this end, we propose a simple yet effective framework, as illustrated in Fig.~\ref{fig:Overview}, to stabilize the weakly-supervised training for GAR. The core idea of our approach is to firstly construct all possible social-representation and then find the effective ones based on the social assumption that {\bf key instances~(people/frames) are closely related to each other.}
	Formally, given a sequence of frames $(V_{1}, V_{2}, \cdots, V_{T})$, our approach models them as follow:
	\begin{align}\label{overall}
	O = \mathcal{O}(\mathcal{F}(V_{1};\mathcal{D}(V_{1});\mathbf{W}), \mathcal{F}(V_{2};\mathcal{D}(V_{2});\mathbf{W}), \cdots,\mathcal{F}(V_{T};\mathcal{D}(V_{T});\mathbf{W})).
	\end{align}{Here, $\mathcal{D}(V_{t})$ represents detecting $N^\mathrm{p}$ proposals from each frame. There are two choices to determine the value of $N^\mathrm{p}$ as follows, i) {\bf Quantity-aware:~\label{Quan-N}}empirically select top-$N^\mathrm{p}$ boxes from numerous proposals; ii) {\bf Probability-aware:~\label{Prob-N}}choose the boxes whose probability is larger than a threshold $\theta$.}
	
	{The spatial modeling function $\mathcal{F}(V_{t};\mathcal{D}(V_{t});\mathbf{W}))$ represents that i) adopt CNN with parameters $\mathbf{W}$ to extract the convolutional feature map for frame $V_{t}$, ii) apply RoIAlign~\cite{he2017mask} to extract person-level features according to the corresponding proposals from $\mathcal{D}(V_{t})$, and iii) fuse person-level features into a single frame-level vector. However, without person-level annotation, it is unavoidable for $\mathcal{D}(\cdot)$ to get many useless proposals from each frame. Moreover, the number of proposals~($N^\mathrm{p}$) varies over samples in practical applications. Thus, $\mathcal{F}(\cdot)$ needs to be able to choose $K^\mathrm{p}$ discriminative person-level features in the spatial domain.} {$\mathcal{O}(\cdot)$ is a temporal modeling function that samples a set of $N^\mathrm{f}$ frames from the entire video sequence~($T$ frames) as the input of our approach according to the sampling strategy used in~\cite{wang2016temporal}. However, the long temporal structure of the activities in our NBA dataset will bring numerous irrelevant frames that may affect the construction of social-representation. Therefore, we also hope $\mathcal{O}(\cdot)$ can select $K^\mathrm{f}$ effective frame-level representations in the temporal domain.}
	
	It is clear that $\mathcal{F(\cdot)}$ and $\mathcal{O(\cdot)}$ need to have similar properties that attending to effective person/frame-level features in the spatial and temporal domain, respectively. Therefore, this work aims at endowing the function $\mathcal{F(\cdot)}$ and $\mathcal{O(\cdot)}$ with the ability of feature selection according to the social assumption that key instances are highly related to each other.

	\subsection{Social Adaptive Module~(SAM)}\label{SAM}
	Inspired by relational reasoning~\cite{santoro2017simple,wang2018non,zhou2018temporal}, we build a generic module, namely Social Adaptive Module, to implement the idea of assisting weakly-supervised training with the social assumption. Specifically, we abstract $\mathcal{F(\cdot)}$ and $\mathcal{O(\cdot)}$ into a unified form as
	\begin{align}\label{SAM_formulation}
	{\mathbf Z} = {\mathcal{M}}({\mathbf X}) &= \{ {\mathbf a} \mid {\mathbf a} \in \{ 
	{\lambda}_{1}  \mathcal{E}({\mathbf x}_{1}), 
	{\lambda}_{2} \mathcal{E}({\mathbf x}_{2}), 
	\cdots, 
	{\lambda}_{N} \mathcal{E}({\mathbf x}_{N}) \},  \notag \\
	& \quad \quad \quad  {\mathbf a}\not = {\mathbf 0}, {\mathbf x}_{i} \in {\mathbf X}, {\lambda}_{i} \in \{ 0, 1\}, {\lVert \boldsymbol \lambda \rVert}_{1}=K
	\},
	\end{align}where ${\mathbf X} \in \mathbb{R}^{N \times D}$ and ${\mathbf Z}\in {\mathbb{R}^{K \times D}}$ are the input and output of $\mathcal{M}(\cdot)$, respectively, and $K\leq N$. Put simply, $\mathcal{M}(\cdot)$ aims to learn the parameter ${\boldsymbol \lambda \in \mathbb{R}^{N} }$, a zero-one vector used to select $K$ discriminative feature from $N$ input feature. $\mathcal{E}(\cdot)$ is the embedding function for input and is optional. We hold that $\boldsymbol{\lambda}$ will be effective for feature selection only if driven by ${\mathbf X}$. Moreover, not only $N \not= K$ but also the value of $N$ varies over samples. Therefore, directly replacing the function $\mathcal{F(\cdot)}$ and $\mathcal{O(\cdot)}$ in Eq.\eqref{overall} with $\mathcal M(\cdot)$ is difficult for our approach to be optimized.
	
	In this work, we approximate the solution of $\boldsymbol{\lambda}$ via pruning a {Dense Relation Graph} with $N$ nodes to a {Sparse Relation Graph} with $K$ nodes via a pruning operation. Specifically, we build a dense relation graph on $N$ input feature to measure relationships between each other. During the process of pruning, we aim at maintaining the top-$K$ feature nodes of the graph according to their relatedness. Based on the $K$ selected features, a sparse relation graph is built to perform relational embedding for them. The details are described as follows.

	{\bf Dense Relation Graph.}~We first build dense relationships between each input node, based on their visual feature. More specifically, given a set of feature vectors as $\{{\mathbf x}_{1}, {\mathbf x}_{2}, \cdots, {\mathbf x}_{N}\}$, we compute the directional relation between them as $r_{ij}=g({\mathbf x}_{i}, {\mathbf x}_{j})$ where $i,j$ are indices and ${g}(\cdot, \cdot)$ is the relation function. There are several common implementations~\cite{liu2019learning,santoro2017simple} of $g(\cdot, \cdot)$. For instance, we can measure the $L_{2}$ distance between each feature, but which is not a data-driven and learnable method. Besides that, we can treat the concatenation $[x_{i}, x_{j}]$ as the input of a multi-layer perceptron to get the relation score. However, as the number of pairs increases, this approach will consume a lot of memory and computation. In this work, we adopt a learnable and low-cost function to measure the relation between $i$-th and $j$-th feature node as $g({\mathbf x}_{i}, {\mathbf x}_{j}) = {\varPhi({\mathbf x}_{i})}^{\top}\varPsi({\mathbf x}_{j}),$ where $\varPhi(\cdot)$ and $\varPsi(\cdot)$ are two embeddings of $i$-th and $j$-th feature node, respectively. {Based on this formulation, the calculation of relation matrices, ${\mathbf R}=\{r_{ij}\}^{N \times N}$, can be implemented by only two embedding processes and a matrix multiplication. We also apply a {\em softmax} computation along the dimension $j$ of the matrix $\mathbf R$.}
	
	{\bf Pruning Operation.}~{To approximate the solution of $\boldsymbol{\lambda}$ in Eq.\eqref{SAM_formulation}, this paper select the $K$ most relevant nodes from the above dense graph based on the social assumption that key instances are likely to be related to each other.} Concretely, after obtaining the $N \times N$ relation matrix for all feature pairs, we construct the {\em relatedness} for each feature node as ${\alpha}_{i} = \sum^{N}_{j=1}{(r_{ij}+r_{ji})},$ where $r_{i*}$ and $r_{*i}$ denote the out-edges and in-edges of $i$-th feature node in the dense relation graph. {Intuitively, the nodes with strong connections can be easily retained in the graph. Thus, we hold that the sum of a specific node’s corresponding connections can depict the importance~({\em relatedness}) of itself.}
	
	Based on the social assumption, we sort the values of ${\boldsymbol \alpha \in \mathbb{R}^{N}}$ in descending order and select the top-$K$ values denoted as ${\mathtt{topk}}(\boldsymbol \alpha) \in \mathbb{R}^{K}$. Thus, the satisfactory $\boldsymbol \lambda$ can be expressed as
	\begin{align}
	{\lambda}_{i} &= 
	\begin{cases}
	1,& {\alpha}_{i} \in {\mathtt{topk}}(\boldsymbol \alpha),\\
	0,& \text{otherwise}.
	\end{cases}
	\end{align}
	
	{\bf Sparse Relation Graph.} According to $\boldsymbol{\lambda}$, we can get the corresponding $K$ selected feature, $\hat{\mathbf X} \in \mathbb{R}^{K \times D}$, namely sparse feature.
	However, $\boldsymbol{\lambda}$ is driven by $\mathbf{R}$, but $\hat{\mathbf X}$ is unrelated to it. Therefore, $\boldsymbol{\lambda}$ will be unlearnable if we directly regard $\hat{\mathbf X}$ as the output of this module. To tackle this problem, we construct a relational embedding $\mathcal{E}(\cdot)$ for the sparse feature $\hat{\mathbf X}$ by combing with relation matrix $\mathbf{R}$. Similarly, we obtain a sparse relation matrix $\hat{\mathbf R}=\{{\hat r}_{ij}\}^{K \times K}$ associating to the $K$ selected feature, and then perform relational embedding as
	\begin{align}
	{\mathbf z}_{i}&= \mathcal{E}(\hat{\mathbf x}_{i}) = {\mathbf W}_{z}\Big({\sum^{K}_{j=1}{\hat{r}_{ij}\varOmega(\hat{\mathbf x}_{j})}}\Big)+ \hat{\mathbf x}_{i}.
	\end{align} Here ``+" denotes a residual connection, $\varOmega(\cdot)$ is the embedding of sparse feature $\hat{\mathbf x}_{j}$, and ${\mathbf W}_{z}$ is a weight vector that projects the relational feature to the new representation with the same dimension as the sparse feature $\hat{\mathbf x}_{i}$. 
	
	{SAM is the first to introduce the “social assumption” that helps a lot in the GAR scenario where many uncertain inputs are involved. More importantly, this makes our method more appropriate to work in the weakly-supervised setting. In comparison:  i)~\cite{yan2018participation} and~\cite{tang2018mining} only built the pair-wise relationship between each player and the scene, but SAM captured the relationship among all people that provides richer information for understanding complex scenes. ii)~\cite{deng2016structure},~\cite{qi2018stagnet},~\cite{tang2019coherence} and~\cite{wu2019learning}, as graph-based methods, indeed built relationships among different people, but they did not provide a mechanism to handle uncertain inputs. Therefore, we believe that SAM can also be used upon these methods.}

	\subsection{Implementation details}
	
	{\bf Person Detection \& Feature Extraction.} For each frame, we first adopt Faster-RCNN \cite{ren2015faster} pre-trained on the MS-COCO~\cite{lin2014microsoft} to detect possible persons in the scene, based on the mmdetection toolbox~\cite{mmdetection}. Then, we track them over all frames by correlation tracker~\cite{danelljan2014accurate} implemented by Dlib~\cite{king2009dlib}. After that, we adopt ResNet-18~\cite{he2016deep} as the backbone to extract the convolutional feature map for each frame. Finally, we get the aligned feature for each proposal from the map by RoIAlign~\cite{he2017mask} with the crop size of $5 \times 5$ and embed it to $1024$ dimensional feature vector by a fully connected layer. 
	
	{\bf Social Adaptive Module.} This module is designed to select out $K$ effective feature from $N$ input ones. However, the values of $N$ and $K$ depend on the situation and will be explained in experiments. If $N$ varies over samples~({\em e.g.}, different numbers of proposals are generated by the Probability-aware strategy mentioned in Section~\ref{Prob-N}), we feed data into this module with a batch-size of 1 but do not change the batch-size of the entire framework. The $\varPhi(\cdot)$, $\varPsi(\cdot)$, and $\varOmega(\cdot)$  used to embed input feature are implemented by $1 \times 1$ convolutional layers. 
	
	{\bf Optimization.} We adopt the ADAM to optimize our approach with fixed hyper-parameters~($\beta_{1}=\beta_{2}=0.9$, $\varepsilon=10^{-4}$) and train it in 30 epochs with an initial learning rate of 0.0001 that is reduced to 1/10 of the previous value for every 5 epochs. Compared with SSU~\cite{bagautdinov2017social} and ARG~\cite{wu2019learning}, which require pre-training the CNN backbone and fine-tuning the top model separately, our approach excluding detection can be optimized in an end-to-end fashion.
	
	\section{Experiments}
	\setlength{\tabcolsep}{4pt}
	\begin{table}[!t]
		\begin{center}
			\caption{Ablation studies on NBA. Quan-$N^\mathrm{p}$ and Prob-$N^\mathrm{p}$ are two different strategies of deciding the number of input proposals, as described in Section~\ref{Overall_Formulation}. $\theta$ is the probability threshold used in Prob-$N^\mathrm{p}$, $N^\mathrm{f}$ is the number of input frames, and $K^{*}$ denote the number of feature selected by our SAM}
			\label{table:NBA_ablation_study}
			\begin{tabular}{llcc}
				\hline\noalign{\smallskip}
				{Type}&Options of Our Approach & Acc~(\%) & Mean Acc~(\%)\\
				\noalign{\smallskip}
				\hline
				\noalign{\smallskip}
				\multirow{4}*{Quan-$N^\mathrm{p}$}&\makecell[l]{B1: w/o SAM~($N^\mathrm{p}=8$)} &44.6 &{39.5}\\
				&\makecell[l]{B2: w/ Spatial-SAM~($N^\mathrm{p}=14, K^\mathrm{p}=14$)}&{46.8} &{41.3}\\
				&\makecell[l]{B3: w/ Spatial-SAM~($N^\mathrm{p}=14, K^\mathrm{p}=8$)}&{\bf 50.3} &{43.6}\\
				&\makecell[l]{B4: w/ Spatial-SAM~($N^\mathrm{p}=8, K^\mathrm{p}=8$)}&{47.4} &{41.4}\\
				\cmidrule(lr){2-4}
				&\makecell[l]{B5: w/ Spatial-SAM~($N^\mathrm{p}=14, K^\mathrm{p}=8$)\\ + w/ Temporal-SAM~($N^\mathrm{f}=20, K^\mathrm{f}=6$)}&{49.1} &{\bf 47.5}\\
				\noalign{\smallskip}
				\hline
				\noalign{\smallskip}
				{Prob-$N^\mathrm{p}$}&\makecell[l]{B6: w/ Spatial-SAM~($\theta=0.9, K^\mathrm{p}=8$)} &{47.5} &{42.6}\\
				\noalign{\smallskip}
				\hline
			\end{tabular}
		\end{center}
	\end{table}
	\setlength{\tabcolsep}{1.4pt}
	
	\subsection{Quantitative Analysis on the NBA Dataset}
	{We first evaluate our approach on the new benchmark by compared with several variants and baseline methods. For this dataset, we sample $N^\mathrm{f}=20$ frames from the entire video clip as the input for all methods and train them with a batch-size of 16. Because of the fast speed of activities in this benchmark, we do not track pre-detected proposals over frames. Moreover, we do not apply any strategy to handle the class-imbalance issue in this benchmark.
	}
	
	{\bf Ablation Study.}~{To evaluate the effectiveness of our SAM, different variants of our approach are performed on NBA, and the results are reported in Table~\ref{table:NBA_ablation_study}. B1 that does not use the proposed SAM achieves the base accuracy of $44.6\%$ and $39.5\%$ on Acc and Mean Acc, respectively. Compared with B1, B2 that employs Spatial-SAM to build relational embedding among $N^\mathrm{p}=14$ proposals but does not prune useless ones, only obtains $2.2\%$ and $1.8\%$ improvement on Acc and Mean Acc. Similarly, B4 which directly adapts Spatial-SAM to generate relation representation from $N^\mathrm{p}=8$ proposals has small improvement. However, by selecting $K^\mathrm{p}=8$ persons from $N^\mathrm{p}=14$ proposals and modeling relationship among them, B3 improves Acc and Mean Acc by $5.7\%$ and $4.1\%$, respectively, compared with B1. Moreover, our Quan-$N^\mathrm{p}$ based approach~(B6) suffering an uncertain number of proposals also gets a satisfactory Mean Acc of $42.6\%$. 
		Based on B3, B5 obtains the best Mean Acc by applying the SAM on the temporal domain. It demonstrates that the ability of feature selection of SAM can also be used to capture the long temporal structure in our NBA dataset.
		The further analysis on the parameters of $N^{*}$ and $K^{*}$ are present in Section~\ref{QA}.}

	{\bf Comparison with the baselines.}~{We also compare our approach with recent work in video classification domain, including TSN~\cite{wang2016temporal}, TRN~\cite{zhou2018temporal}, I3D~\cite{carreira2017quo}, I3D+NLN~\cite{wang2018non}. To be fair, all these baseline methods are built on ResNet-18 and the input modality is RGB. The results are reported in Table~\ref{table:baseline_NBA}. We see that ``Ours w/o SAM" is hardly improved or worse due to noise input~(irrelevant pre-detected proposals), compared with methods~(``TSN" and ``TRN") only using frame-level information. By introducing SAM to select discriminative proposals in the spatial domain, ``Ours w/ SAM~(S)" achieves significant improvement on Mean Acc but still overfits on some classes. As expected, ``Ours w/ SAM~(S+T)" outperforms all baselines by a good margin and obtained the best Mean Acc by simultaneously applying SAM to the spatial and temporal domain. Nevertheless, ``Ours w/ SAM~(S+T)" performs poorly on the activity of ``3p-succ." which does not have long-term temporal structure. Moreover, ``I3D" and ``I3D+NLN" which depend on dense frames perform poorly on this benchmark. 
	}

	\setlength{\tabcolsep}{4pt}
	\begin{table}[!t]
		\begin{center}
			\caption{Comparison on NBA. ``Ours w/o SAM", ``Ours w/ SAM~(S)", and ``Ours w/ SAM~(S+T)" are the B1, B3, and B5 reported in Table~\ref{table:NBA_ablation_study}, respectively}
			\label{table:baseline_NBA}
			\begin{tabular}{lccccccc}
				\hline\noalign{\smallskip}
				\multirow{3}*{\makecell[l]{Group Activity}}& \multicolumn{4}{c}{Frame Classification}&\multicolumn{3}{c}{Our Approach}\\
				\cmidrule(lr){2-5}\cmidrule(lr){6-8}
				&{\makecell[c]{TSN\\\cite{wang2016temporal}}} &{\makecell[c]{TRN\\\cite{zhou2018temporal}}} &{\makecell[c]{I3D\\\cite{carreira2017quo}}} &{\makecell[c]{I3D+NLN\\\cite{wang2018non}}}&w/o SAM & \makecell[c]{w/ SAM\\(S)}  & \makecell[c]{w/ SAM\\(S+T)}\\
				\noalign{\smallskip}
				\hline
				\noalign{\smallskip}
				2p-succ.                         &38.7              &{44.8}            & 33.1            &22.1            & {46.6}             &{39.3}             &{\bf47.2}\\
				2p-fail.-off.                      &{30.8}             &{23.4}       & 14.0             &20.6        & 28.0                   &25.2                 &{\bf42.1}\\
				2p-fail.-def.                     &49.1              &50.0            & 39.3            &45.3            & 49.6                    &{\bf 53.4}       &48.3\\
				2p-layup-succ.                 &52.9             & 54.7             & 50.6            &48.8            & 44.2                 &{\bf 57.6}         &53.5\\
				2p-layup-fail.-off.             &10.1             & {22.5}        & {22.5}     &{22.5}            & {20.2}      &{19.1}                   &{\bf32.6}\\
				2p-layup-fail.-def.             &44.6             & 46.5            & 43.3            &31.2            & 44.6                  &{51.6}          &{\bf59.9}\\
				3p-succ.                          &39.3            &37.7            &31.1             &26.8            & 39.9                      &{\bf41.0}           &30.1\\
				3p-fail.-off.                     &10.8            & 20.5              & 4.8              &12.0            & {24.1}                &{38.6}          &{\bf55.4}\\
				3p-fail.-def.                     &63.9             & 62.8             & 55.3            &61.7            &{58.6}                 &{\bf66.9}            &58.1\\
				\noalign{\smallskip}
				\hline
				\noalign{\smallskip}
				Mean Acc~(\%)                 &37.8             &40.3            &32.7        &32.3        &39.5                 &{43.6}           &{\bf47.5} \\
				\noalign{\smallskip}
				\hline
			\end{tabular}
		\end{center}
	\end{table}
	\setlength{\tabcolsep}{1.4pt}

	\subsection{Qualitative Analysis on the NBA dataset}\label{QA}
\begin{figure}[!t]
		\centering
		\subfigure[$N^\mathrm{p}$ of Spatial-SAM]{\includegraphics[width=0.32\textwidth]{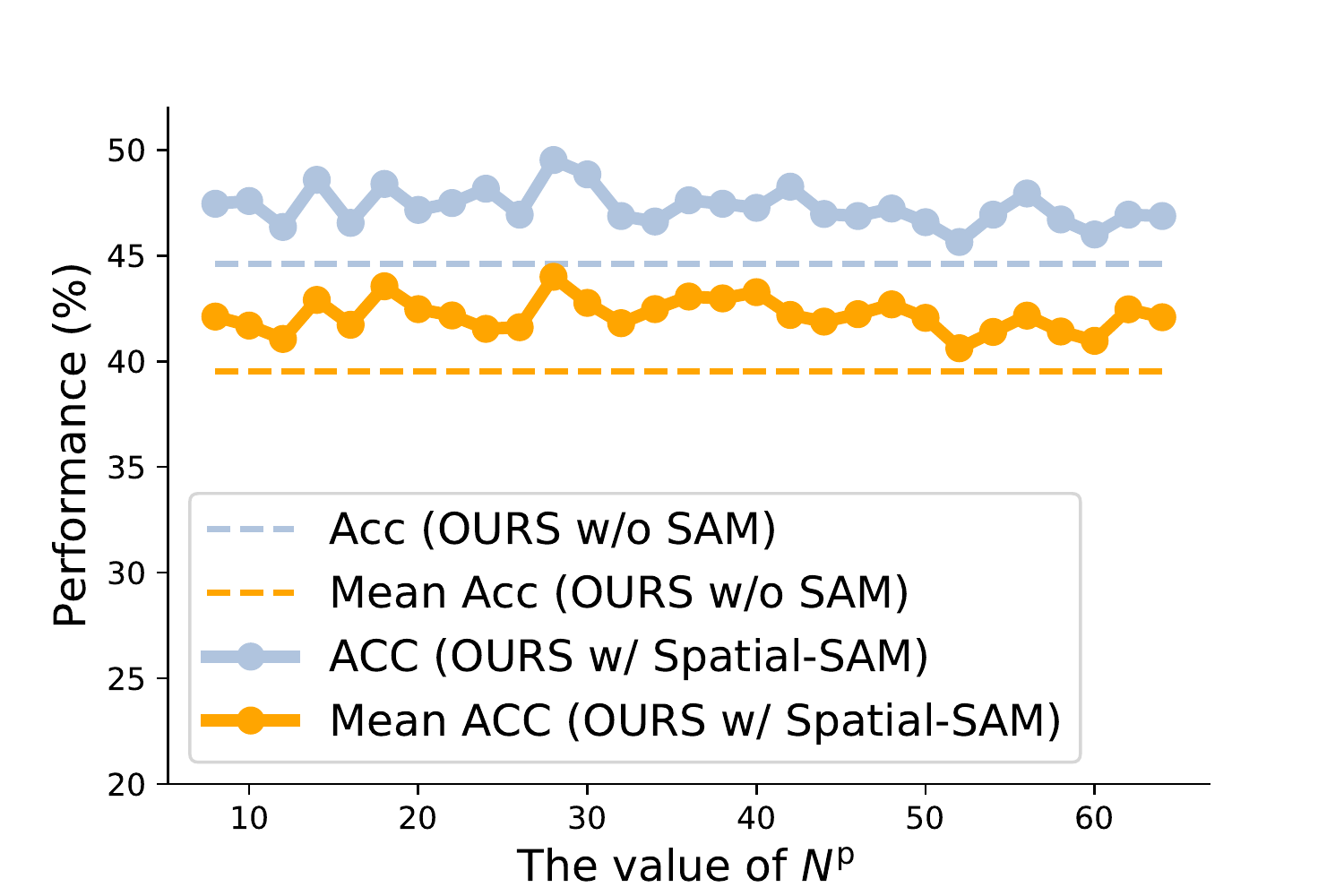}}
		\subfigure[$\theta$ of Spatial-SAM]{\includegraphics[width=0.32\textwidth]{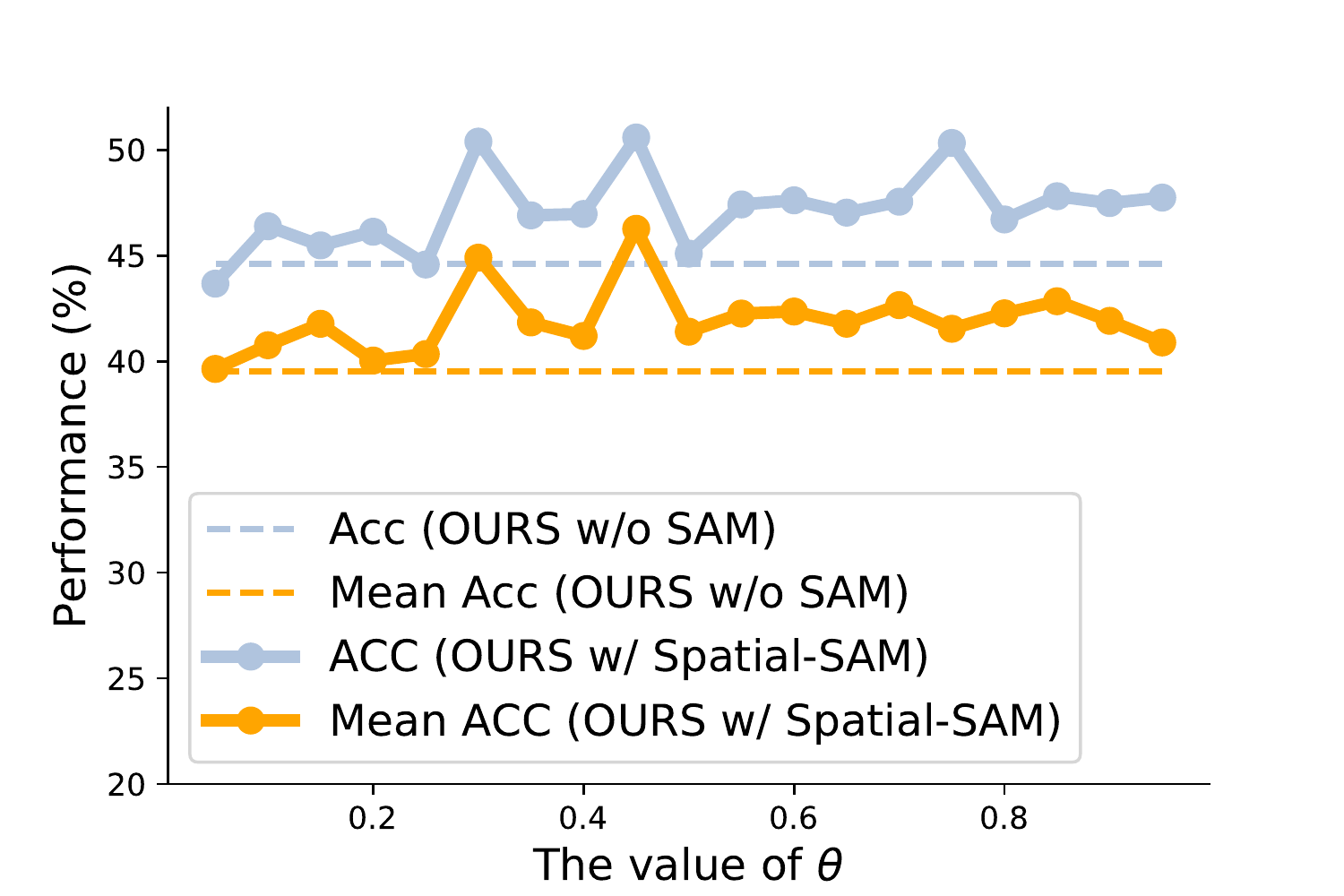}}
		\subfigure[$K^\mathrm{p}$ of Spatial-SAM]{\includegraphics[width=0.32\textwidth]{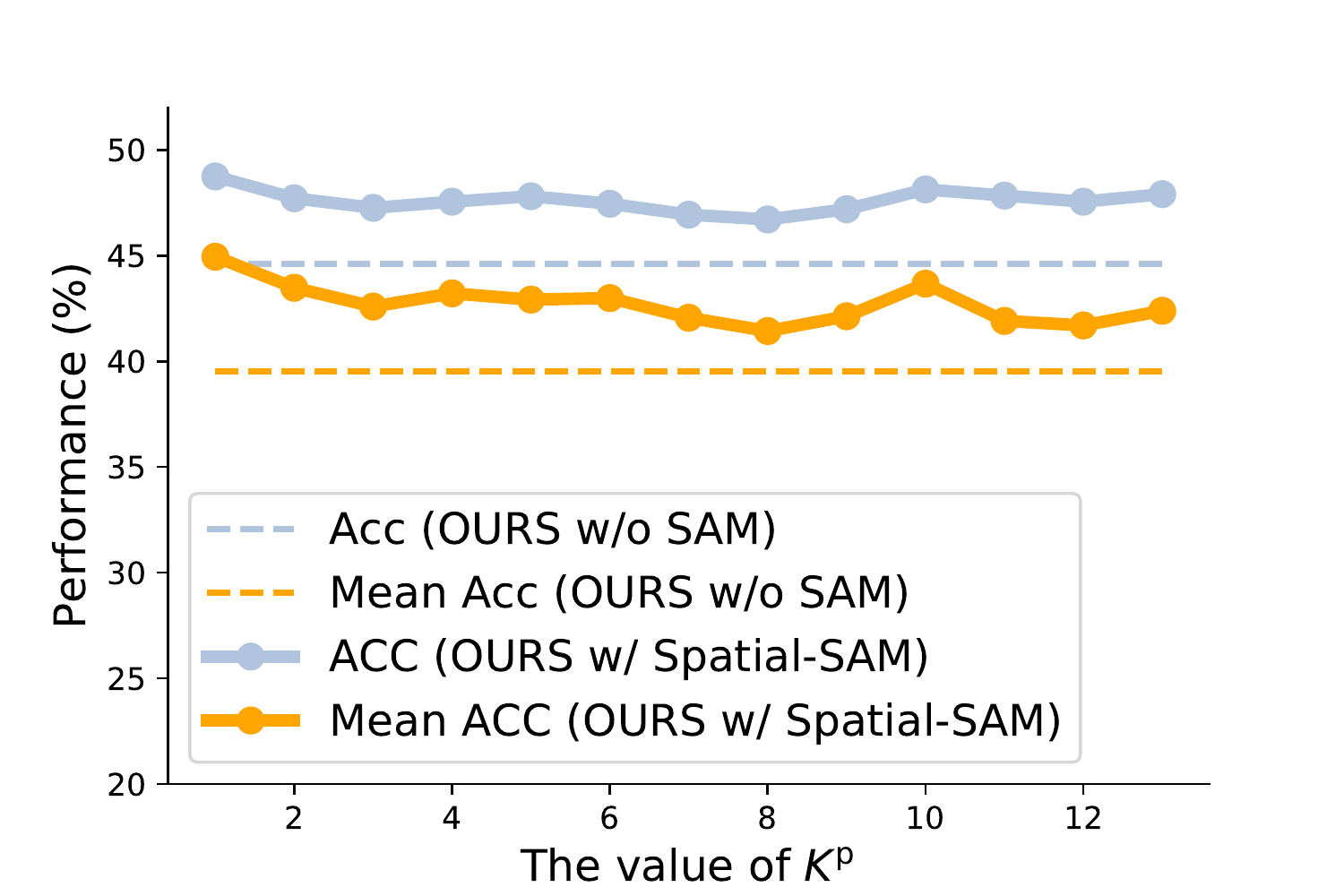}}
		\subfigure[$K^\mathrm{f}$ of Temporal-SAM]{\includegraphics[width=0.32\textwidth]{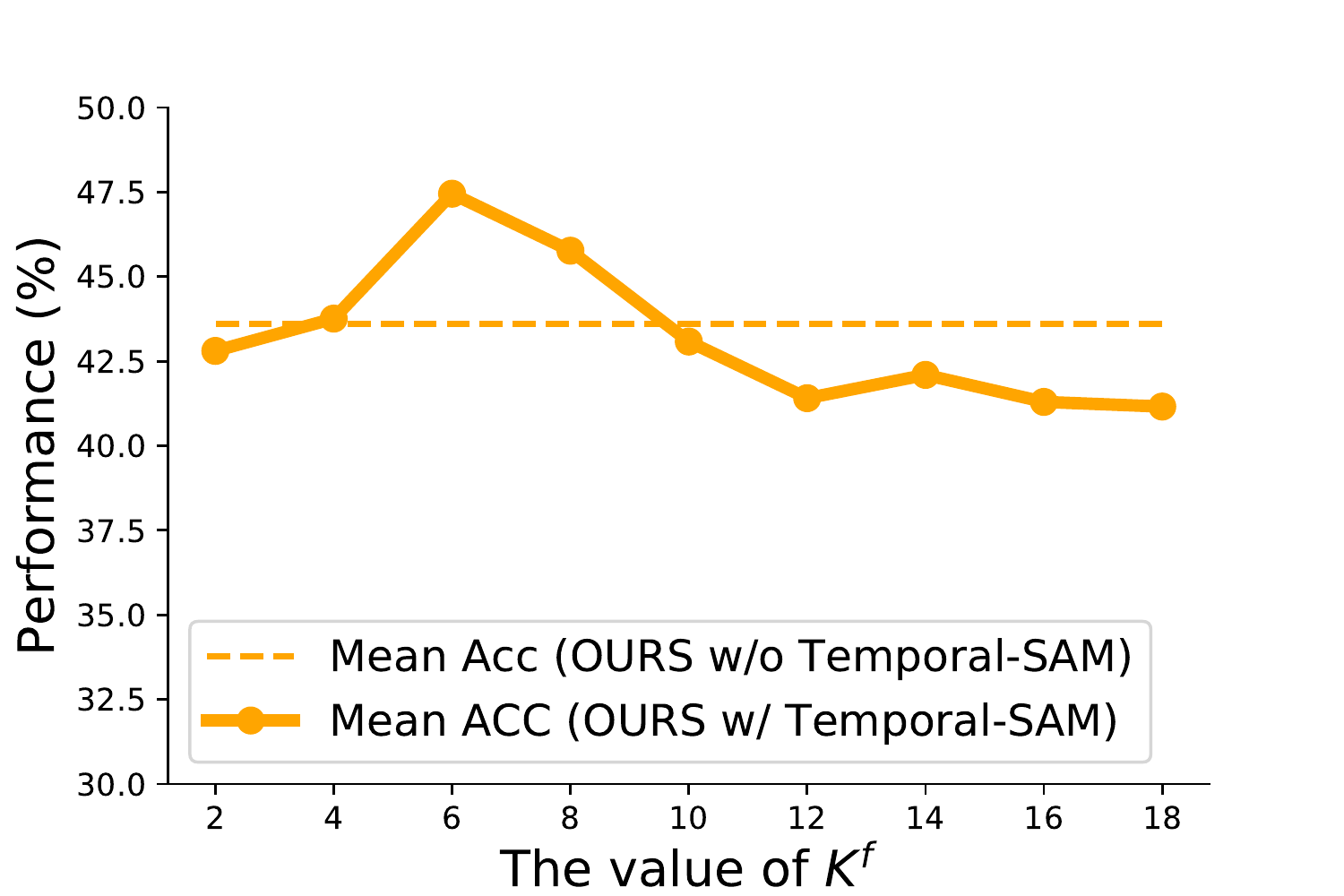}}
		\subfigure[{Confusion matrix}]{\includegraphics[width=0.32\textwidth]{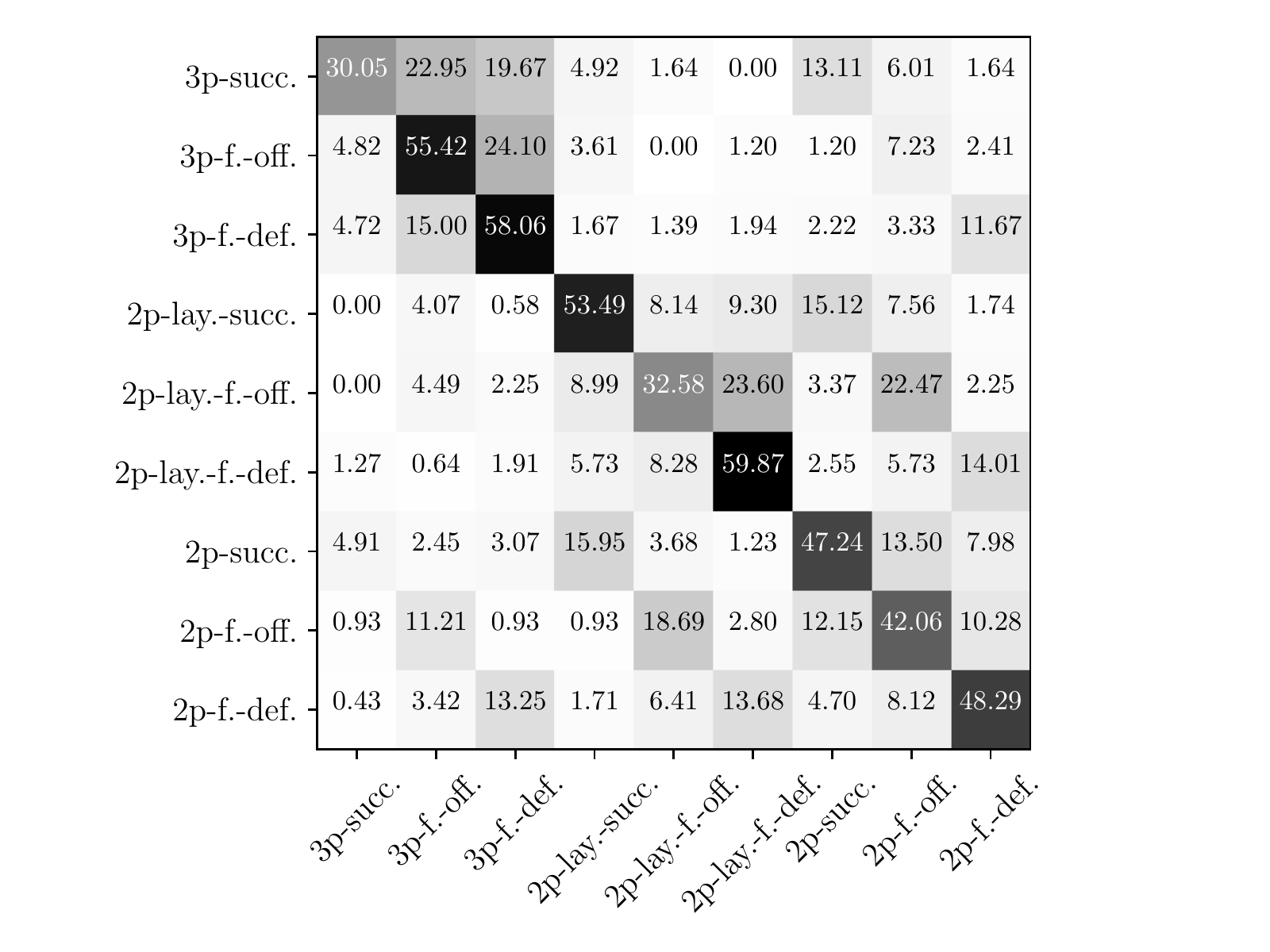}}
		\subfigure[Embeddings of ``shot"]{\includegraphics[width=0.32\textwidth]{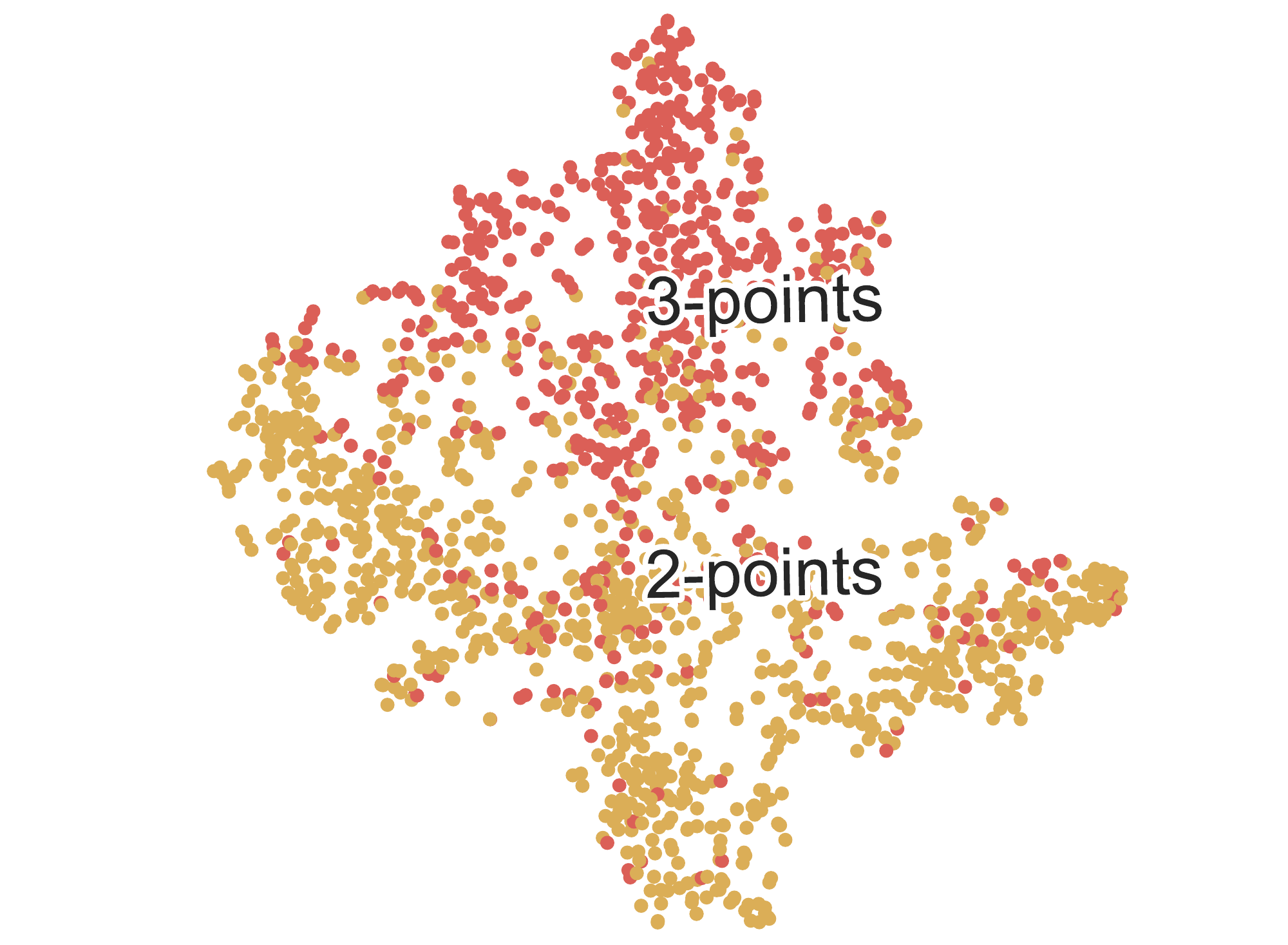}}
		\caption{(a)-(d)~Experimental analysis on parameters. (e)~The confusion matrix of OURS w/ Spatial-SAM and Temporal-SAM. (f)~t-SNE visualization of embeddings of 2/3-points based activities. These experiments are carried out on the NBA dataset}
		\label{fig:further_analysis}
	\end{figure}

	{{\bf Analysis of parameters.} We first diagnose $N$, the number of nodes of the dense relation graph. Limited by the computation resource, we only analyze the $N^\mathrm{p}$ of Spatial-SAM and it indicates how many pre-detected proposals should be fed into our approach. It can be decided by two strategies as mentioned in Section~\ref{Overall_Formulation}. Thus, we first run our Quan-$N^\mathrm{p}$ based approach on the NBA dataset by fixing $K^\mathrm{p}=8$ and changing $N^\mathrm{p}$ from $8$ to $64$ with a step of $4$. As shown in Fig.~\ref{fig:further_analysis}(a), although $N^\mathrm{p}$ is increasing, the performance of our approach has been persistently higher than the baseline. Moreover, we also conduct our Prob-$N^\mathrm{p}$ based approach on NBA by using fixed $K^\mathrm{p}=8$ and adjust $\theta$ from $0.05$ to $0.95$ with a mini-step of $0.05$. As shown in Fig.~\ref{fig:further_analysis}(b), our approach can achieve promising results when $\theta \geq 0.3$ and is more likely to get high performance when $\theta$ around $0.4$. Overall, our Spatial-SAM is not sensitive to $N^\mathrm{p}$ whether decided by Quan-$N^\mathrm{p}$ or Prob-$N^\mathrm{p}$.}
	{We also diagnose $K$, the number of nodes of the sparse relation graph, and it decides how many feature nodes need to be selected for modeling. As shown in Fig.~\ref{fig:further_analysis}(c), the performance of Spatial-SAM maintains over the baseline and it obtains the best result at $K^\mathrm{p}=1$. Therefore, we hold that Spatial-SAM is not sensitive to $K^\mathrm{p}$. By contrast, the performance of Temporal-SAM cannot get satisfactory performance when the $K^\mathrm{f}$ is too small or large, due to the different temporal length of activities in NBA. However, our approach with Temporal-SAM significantly improves Mean Acc when $4< K^\mathrm{f} < 10$.}
	
	{\bf Confusion matrix.} To figure out the confusion between each activity in the NBA dataset, we report the confusion matrix of our approach in Fig.\ref{fig:further_analysis}(d). We can see that the activities involving ``defense" and ``offense" are easily confused, due to the class-imbalance issue between these two kinds of activities. However, it is relatively easy to distinguish 2-points and 3-points, as embeddings shown in Fig.\ref{fig:further_analysis}(f). Because 3-point players usually jump to shot behind the 3-point line without blocking. By contrast, 2-point players are often blocked by others.
	
	\begin{figure}[h]
		\centering
		{\includegraphics[width=0.95\textwidth]{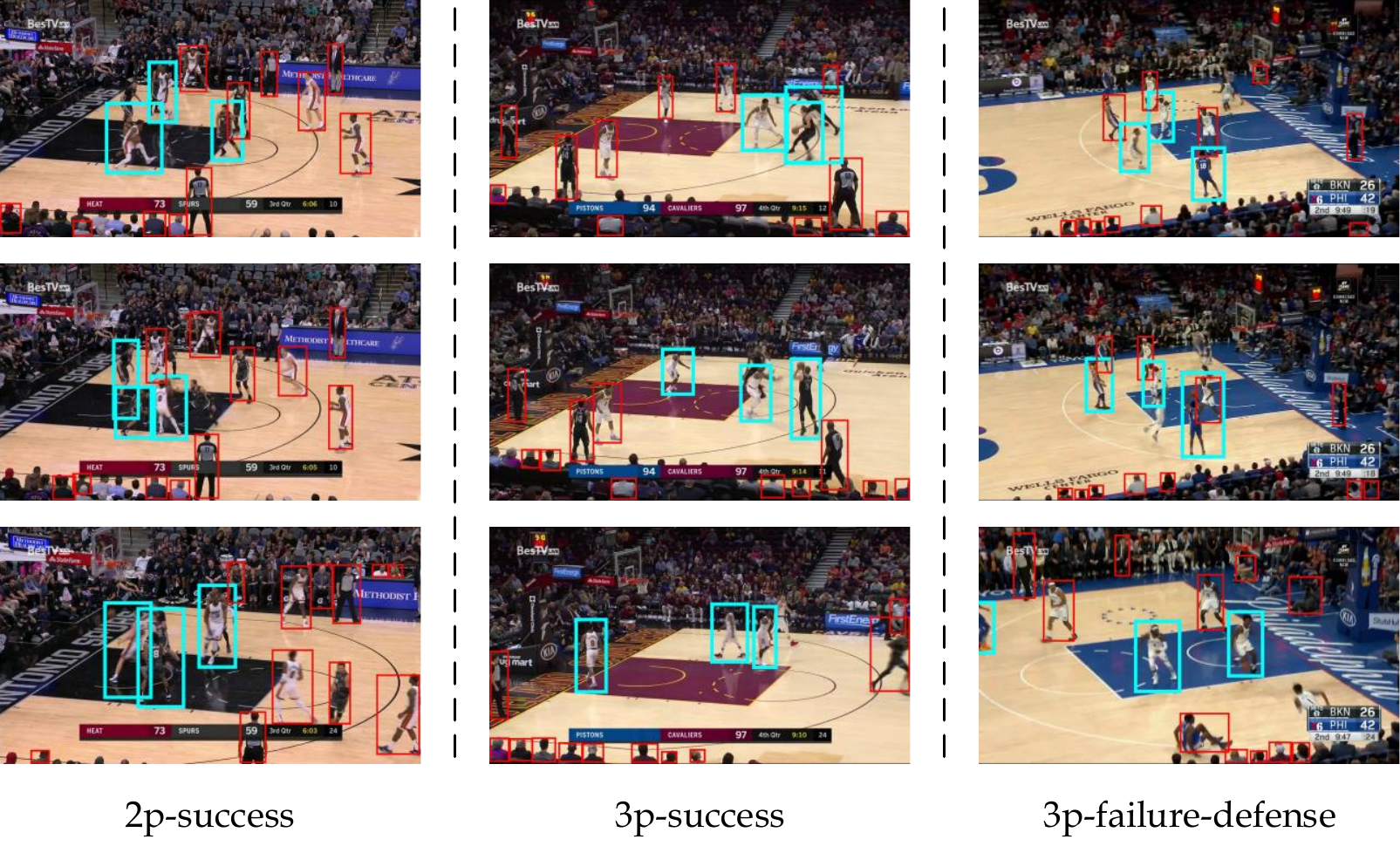}}
		\caption{A visualization of the top-$K$ proposals focused by SAM over time on the NBA dataset, where $K=3$. Each column shows three different frames of an activity. We highlight the top-$K$ players~(in {\color{cyan}cyan} boxes) at three time steps of different activities. The people in {\color{red}red} boxes are treat as noisy data by our model}
		\label{fig:BD_case}
	\end{figure}
	
	{
		{\bf Visualization.} To further understand the discriminative learning process of SAM, we show some typical cases of NBA in Fig.\ref{fig:BD_case}. The group activities in NBA have long-term temporal information, thus top-$K$ proposals vary over time. {Take the rightmost one as an example, a “3p-failure-defense” has 3 parts: (1) preparation, (2) shooting, (3) defensive rebound. For (1) and (2), the players controlling the ball are the key instances, but for (3), the players that quickly turn back are the key instances}. It is not hard to find that SAM aims at focusing on the players who are controlling the basketball or close to it and these people can form a group semantically.
	}

	\subsection{Quantitative Analysis on Volleyball Dataset}
	We also evaluate our approach on the existing largest and most common benchmark, Volleyball Dataset (VD)~\cite{ibrahim2016hierarchical}
	consisting of 4830 volleyball game sequences. The middle frame of each sequence is labeled with 9 action labels (not used in our approach) and 8 group activity labels.
	However, we find that there are many wrong annotations between ``pass" and ``set", which seriously affects the evaluation for models, thus we merged them into ``pass-set". To be fair, we follow the train/test split provided in~\cite{ibrahim2016hierarchical} and sample $N^\mathrm{f}=3$ from the video clip similar to~\cite{wu2019learning}. Because the activities in VD always occur in the middle frame, we do not apply our SAM to the temporal domain for this benchmark.

	{\bf Ablation Study.}~We also perform ablation study on VD and the experimental results are reported in Table~\ref{table:VD_results}(a). 
	All these variants do not use person-level supervision information~(bounding boxes and action labels) provided by~\cite{ibrahim2016hierarchical} and are built on ResNet-18. Compared with the baseline method B1, our B2 and B3 which only apply SAM to generate relational embedding for proposals but do not prune the irrelevant ones, only improve the accuracy by $0.9\%$ and $0.4\%$, respectively. Besides, by using SAM to build relationships among $N=16$ proposals and choosing $K=12$ effective proposals from them, B3 and B5 improve the accuracy of $1.6\%$ based on whether Quan-$N$ or Prob-$N$. This observation indicates again that useless proposals will affect the weakly-supervised training and SAM is effective for pruning them.
	
	{{\bf Comparison with the state-of-the-art.} Referring to~\cite{wang2017recurrent}, we report the results of HTDM~\cite{ibrahim2016hierarchical,msibrahiPAMI16deepactivity}, PCTDM~\cite{yan2018participation}, CCGL~\cite{tang2019coherence}, and StagNet~\cite{qi2018stagnet} by computing their corresponding confusion matrices. We reproduce the state-of-the-art method, ARG~\cite{wu2019learning}, with fully-supervised and weakly-supervised settings, respectively. As shown in Table~\ref{table:VD_results}(b), our weakly-supervised approach with the backbone of ResNet-18 is superior to almost all previous fully-supervised methods, except ARG which is built on Inception-v3. But our approach goes far beyond ARG under the weakly-supervised setting, suggesting that useless pre-detected proposals seriously affect the construction of relation graphs in ARG. Furthermore, our approach with Inception-v3 can achieve the best performance.}
	
	\setlength{\tabcolsep}{4pt}
	\begin{table}[!t]
		\begin{center}
			\caption{Results on VD. (a) Ablation studies. (b) Comparison with SOTA. ``Ours" represents ``Ours w/ Spatial-SAM" with $N^\mathrm{p}=16$ and $K^\mathrm{p}=12$ based on Quan-$N^\mathrm{p}$ }
			\label{table:VD_results}
			\begin{minipage}[t]{0.54\textwidth}
				\centerline{(a)}
				{\smallskip}
				\begin{tabular}{llc}
					\hline\noalign{\smallskip}
					Type& Our Approach  &Acc~(\%) \\
					\noalign{\smallskip}
					\hline
					\noalign{\smallskip}
					\multirow{8}*{Quan-$N^\mathrm{p}$}&\makecell[l]{B1: w/o SAM} &{91.5} \\
					\cmidrule(lr){2-3}
					&\makecell[l]{B2: w/ Spatial-SAM\\($N^\mathrm{p}$=16, $K^\mathrm{p}$=16)}&{92.4} \\
					\cmidrule(lr){2-3}
					&\makecell[l]{B3: w/ Spatial-SAM\\($N^\mathrm{p}$=16, $K^\mathrm{p}$=12)}&{\bf 93.1} \\
					\cmidrule(lr){2-3}
					&\makecell[l]{B4: w/ Spatial-SAM\\($N^\mathrm{p}$=12, $K^\mathrm{p}$=12)}&{91.9} \\
					\noalign{\smallskip}
					\hline
					\noalign{\smallskip}
					\multirow{1}*{Prob-$N^\mathrm{p}$}&\makecell[l]{B5: w/ Spatial-SAM\\($\theta = 0.9, K^\mathrm{p}=12$)}&{\bf 93.1} \\
					\noalign{\smallskip}
					\hline
				\end{tabular}
			\end{minipage}
			\hfil
			\begin{minipage}[t]{0.4\textwidth}
				\centerline{(b)}
				{\smallskip}
				\begin{tabular}{lcc}
					\hline\noalign{\smallskip}
					{Method}  &Supervision &Acc~(\%) \\ 
					\noalign{\smallskip}
					\hline
					\noalign{\smallskip}
					HTDM   &Fully & 89.7  \\ 
					PCTDM &Fully & 90.2 \\ 
					CCGL &Fully& 91.0  \\ 
					StagNet  &Fully& 90.0  \\ 
					{$^{\ddagger}$ARG}  &Fully& {\bf 94.0}  \\ 
					{$^{\ddagger}$ARG} &Weakly& 90.7  \\ 
					\noalign{\smallskip}
					\hline
					\noalign{\smallskip}
					{$^{\dagger}$Ours}&Weakly& {93.1} \\ 
					{$^{\ddagger}$Ours}&Weakly& {\bf 94.0} \\ 
					\noalign{\smallskip}
					\hline
					\noalign{\smallskip}
					\noalign{\smallskip}
					\multicolumn{2}{l}{$^{\dagger}$ ResNet-18}\\
					\multicolumn{2}{l}{$^{\ddagger}$ Inception-v3}\\
				\end{tabular}
			\end{minipage}
		\end{center}
	\end{table}
	\setlength{\tabcolsep}{1.4pt}
	
	\section{Conclusions}
	In this work, we introduce a weakly-supervised setting for GAR, which is more practical and friendly for real-world scenarios. To investigate this problem, we collect a larger and more challenging dataset from high-resolution basketball videos of NBA. Furthermore, we propose a social adaptive module~(SAM) for assisting the weakly-supervised training by leveraging the social assumption that discriminative features are highly related to each other. SAM can be easily plugged into existing frameworks and be optimized in an end-to-end fashion. As demonstrated on two datasets, our approach achieves state-of-the-art results while it can attend to key proposals/frames automatically. 
	
	This work reveals that social relationship among visual entities is helpful for high-level semantic understanding. We look forward to applying this method to more challenging scenarios, in particular, for mining semantic knowledge from weakly-annotated or un-annotated visual data.
	
	\section*{Acknowledgements}
	
	This work was supported by the National Key Research and Development Program of China under Grant 2018AAA0102002, the National Natural Science Foundation of China under Grants 61732007,  61702265, and 61932020.
	
	%
	%
	\bibliographystyle{splncs04}
	\bibliography{eccv}
\end{document}